\documentclass[10pt,twocolumn,letterpaper]{article}

\usepackage{cvpr}              %

\usepackage{amsfonts}
\usepackage{amsmath}
\usepackage[bb=dsserif]{mathalpha}
\usepackage{bm}
\usepackage{booktabs}
\usepackage{graphicx}
\usepackage[pagebackref=true,breaklinks=true,letterpaper=true,colorlinks,bookmarks=false]{hyperref}
\usepackage{siunitx}

\newcommand{\linformermhsamflops}{\num{246.73}}
\newcommand{\sparsifinermhsamflops}{\num{224.04}}
\newcommand{\sparsifinerlinformerimprovement}{\num{2.1}}

\def\cvprPaperID{12118} %
\def\confName{CVPR}
\def\confYear{2023}

\newcommand{\myparagraph}[1]{\textbf{#1 ---}}

\begin{document}

\title{Sparsifiner\@: Learning Sparse Instance-Dependent Attention \\ for Efficient Vision Transformers}

\author{%
{\fontsize{11}{12}\selectfont Cong Wei\textsuperscript{1}\textsuperscript{\thanks{Equal contribution.}}}
\hspace{1mm}
{\fontsize{11}{12}\selectfont Brendan Duke\textsuperscript{1,3}\textsuperscript{\footnotemark[1]}}
\hspace{1mm}
{\fontsize{11}{12}\selectfont Ruowei Jiang\textsuperscript{3}}
\hspace{1mm}
{\fontsize{11}{12}\selectfont Parham Aarabi\textsuperscript{1,3}}
\hspace{1mm}
{\fontsize{11}{12}\selectfont Graham W.~Taylor\textsuperscript{2,4}}
\hspace{1mm}
{\fontsize{11}{12}\selectfont Florian Shkurti\textsuperscript{1,4}}
\vspace{0.1mm}
\and
{\fontsize{10.5}{12}\selectfont \textsuperscript{1}University of Toronto}\and
{\fontsize{10.5}{12}\selectfont \textsuperscript{2}University of Guelph}\and
{\fontsize{10.5}{12}\selectfont \textsuperscript{3}Modiface, Inc}.\and
{\fontsize{10.5}{12}\selectfont \textsuperscript{4}Vector Institute}
}

\maketitle
\pagenumbering{gobble}

\begin{abstract}
	\noindent
	Vision Transformers (ViT) have shown their competitive advantages performance-wise compared to convolutional neural networks (CNNs) though they often come with high computational costs. To this end, previous methods explore different attention patterns by limiting a fixed number of spatially nearby tokens to accelerate the ViT's multi-head self-attention (MHSA)
	operations.
	However, such structured attention patterns limit the token-to-token connections to their spatial relevance, which disregards learned semantic connections from a full attention mask.
	In this work, we propose a novel approach to learn instance-dependent attention patterns, by devising a lightweight connectivity predictor module to estimate the
	connectivity score of each pair of tokens.
	Intuitively, two tokens have high connectivity scores if the features are considered relevant either spatially or semantically. As each token only attends to a small number of other tokens, the binarized connectivity masks are often very sparse by nature and therefore provide the opportunity to accelerate the network via sparse computations.
	Equipped with the learned unstructured attention pattern, sparse attention ViT (Sparsifiner)
	produces a superior Pareto-optimal trade-off between FLOPs and top-1 accuracy on
	ImageNet compared to token sparsity. Our method reduces
	48\% $\sim$ 69\% FLOPs of MHSA while the accuracy drop is within 0.4\%.
	We also show that combining attention and token sparsity reduces ViT FLOPs by over 60\%.
\end{abstract}

\section{Introduction}

\begin{figure}[t]
	\includegraphics[width=\columnwidth]{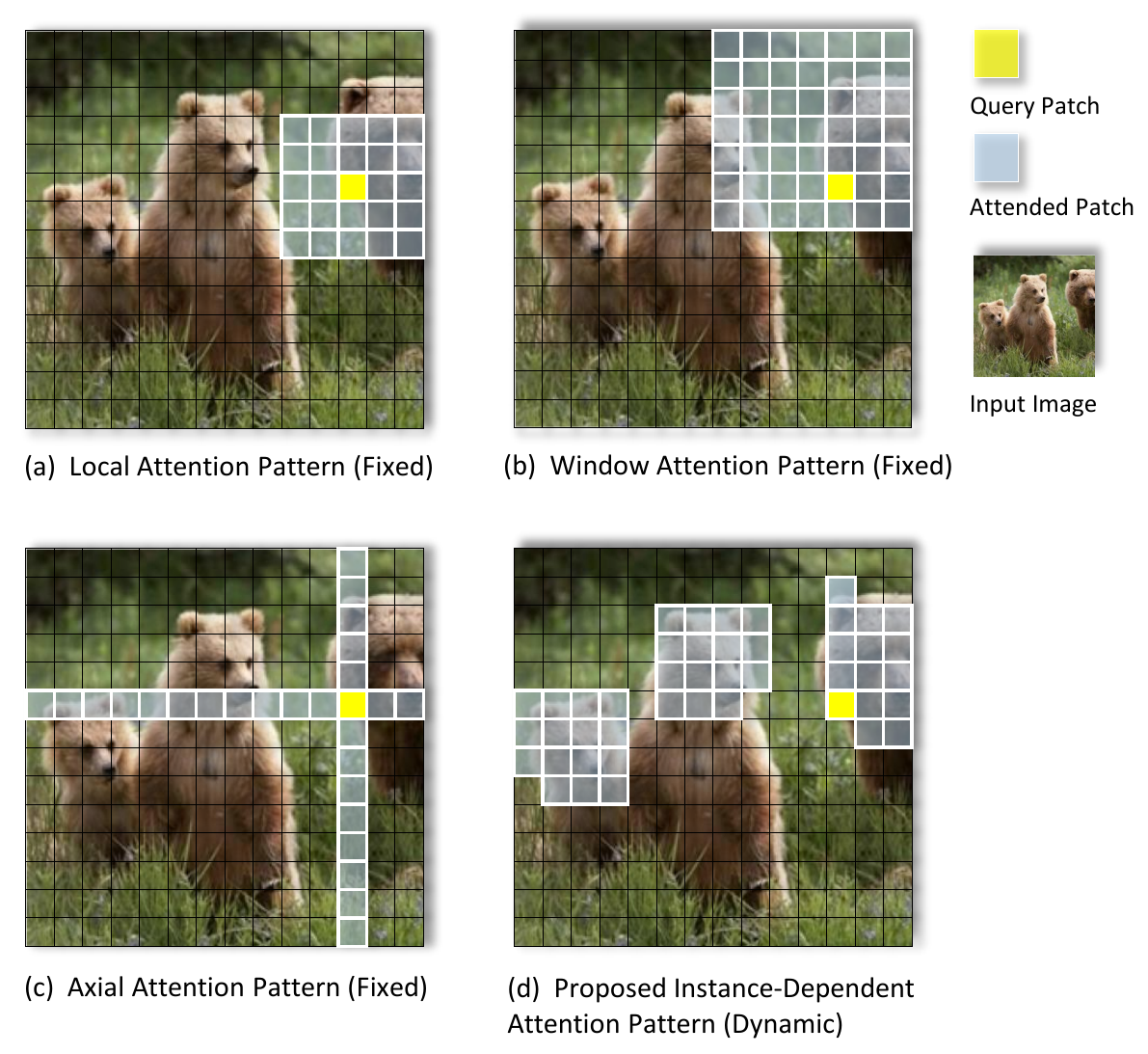}
	\caption{\label{fig:attention-pattern-comparision}Comparison of Sparsifiner and fixed attention patterns.
		Twins~\cite{chu2021twins} (a), Swin~\cite{liu2021swin} (b), and Axial~\cite{ho2019axial} (c)
		address quadratic MHSA complexity using fixed attention patterns,
		which does not consider the instance-dependent nature of semantic information in images.
		To address this, we propose Sparsifiner (d): an efficient module for sparse
		instance-dependent attention pattern prediction.}
\end{figure}
\noindent
Vision Transformers (ViTs)~\cite{dosovitskiy2020animage} have emerged as a dominant
model for fundamental vision tasks such as image classification~\cite{dosovitskiy2020animage},
object detection~\cite{carion2020detr}, and semantic
segmentation~\cite{cheng2021maskformer,cheng2021mask2former}.
However, scaling ViTs to a large number of tokens is challenging due to the
quadradic computational complexity of multi-head self-attention
(MHSA)~\cite{vaswani2017attention}.
This is particularly disadvantageous for large-scale vision tasks because
computing on high-resolution and high-dimensionality inputs is desirable.
For example, input modalities such as video frames and 3D point clouds have a
large number of tokens even for basic use cases.
Novel algorithms are needed to continue to scale ViTs to larger, more complex
vision tasks.

Prior works have largely taken
two approaches to improve the computational efficiency of ViTs: token pruning and using fixed sparse attention patterns in
MHSA\@.
Token pruning methods~\cite{rao2021dynamicvit} reduce the number of tokens by a
fixed ratio called the keep rate, but accuracy degrades quickly when pruning
early layers in the
network~\cite{feng2022evit,ryoo2021tokenlearner,tang2022patchslimming}.
For example, introducing token pruning into shallower layers of
EViT~\cite{feng2022evit} causes a significant~\num{3.16}\% top-1 accuracy drop
on ImageNet~\cite{deng2009imagenet}.
This issue is due to the restriction of pruning an entire token, which
account to pruning an entire row and column of the attention matrix at once.
One way to alleviate this is to prune individual connectivities of the
attention matrix instead of entire tokens.
Existing methods that take this attention matrix connectivity-pruning approach
use fixed sparse attention patterns~\cite{child2019sparsetransformer}.
For example, local and strided fixed attention patterns are
used~\cite{child2019sparsetransformer,duke2021sstvos}, in combination with
randomly-initialized connectivities~\cite{zaheer2020bigbird}.
However, such fixed attention patterns limit the the capacity of the
self-attention connections to a fixed subset of tokens
(Fig.~\ref{fig:attention-pattern-comparision}).
These attention patterns' fixed nature is less effective compared with the
direct communication between tokens in full self attention.
For example, Swin transformer~\cite{liu2021swin,liu2021swinv2} has a limited
receptive field at shallower layers and needs many layers to model long-range
dependencies.
And BigBird~\cite{zaheer2020bigbird} needs to combine multiple fixed attention
patterns to achieve good performance.
Rather, it is desirable to design sparse attention algorithms that mimic full
self attention's instance-dependent nature~\cite{vaswani2017attention}, thereby
capturing the variable distribution of semantic information in the input image
content.

To address aforementioned challenges, we propose a method called Sparsifiner
that learns to compute sparse connectivity patterns over attention that are
both instance-dependent and unstructured.
The instance-dependent nature of the attention pattern allows each token to use
its limited attention budget of nonzero elements more efficiently compared to
fixed sparse attention patterns.
For example, in attention heads that attend to semantic rather than positional
content~\cite{vaswani2017attention,voita2019analyzing}, tokens containing
similar semantic information should be considered to have high connectivity
scores despite their spatial distance.
Similarly, nearby tokens with irrelevant semantic relation should have lower
connectivity scores despite their spatial proximity.
Furthermore, Sparsifiner improves attention pattern flexibility compared to
token pruning by pruning individual connectivities, instead of entire rows and
columns of the attention matrix.
This allows Sparsifiner to reduce FLOPs in the early layers of the network
without incurring significant top-1 accuracy degradation
(\S\ref{sec:experiments}).
By pruning individual connectivities dependent on image content, Sparsifiner
generalizes prior approaches to sparsifying MHSA in ViTs, and in doing so
produces a favourable trade-off between accuracy and FLOPs.

Our contributions can be summarized as:
\begin{itemize}
	\item We propose a novel efficient algorithm called Sparsifiner to predict
	      instance-dependent sparse attention patterns using low-rank connectivity
	      patterns.
	      Our investigation into instance-dependent unstructured sparsity is to
	      the best of our knowledge novel in the context of ViTs.
	\item We show that such learned unstructured attention sparsity produces a superior
	      Pareto-optimal tradeoff between FLOPs and top-1 accuracy on ImageNet compared
	      to token sparsity.
	      Furthermore, we show that Sparsifiner is complementary to token
	      sparsity methods, and the two approaches can be combined to achieve
	      superior performance-accuracy tradeoffs.
	\item We propose a knowledge distillation-based approach for training
	      Sparsifiner from pretrained ViTs using a small number of training epochs.
\end{itemize}

\section{Related Work}%
\label{sec:related}

\noindent
\myparagraph{Efficient Attention}
Developing an efficient attention mechanism for high resolution image encoding is the focus of this work.
Efficient attention mechanisms have been widely studied in NLP tasks to model long sequences.
They can be categorized as follows:
\textbf{Low-rank methods} such as Linformer~\cite{wang2020linformer} use a low-rank projection to linearize the multi-head attention operation.
Linformer~\cite{wang2020linformer} replaces the scaled dot product with linear attention that approximates the attention with a low-rank matrix.
\textbf{Kernelization}, including Performer~\cite{choromanski2021rethinking},
Linear Transformers~\cite{katharopoulos2020transformers}, and Random Feature
Attention~\cite{peng2021random} use kernels to avoid explicitly computing the attention matrix.
\textbf{Sparse attention with fixed attention patterns}~\cite{child2019sparsetransformer,ho2019axial,parmar2018image,qiu-etal-2020-blockwise,chu2021twins}.
This type of technique sparsifies the attention matrix by limiting the field of view to predefined patterns such as local and strided windows.
\textbf{Similarity and clustering-based methods} including Routing Transformer~\cite{roy2021efficient}, Reformer~\cite{Kitaev2020Reformer:}, and Sinkhorn Transformer~\cite{tay2020sparse}.
These models measure token relevance by sorting or clustering and then assign tokens to buckets for within-bucket attention.
\textbf{Neural memory mechanisms} such as Set Transformer~\cite{lee2019set}, Compressive Transformer~\cite{Rae2020Compressive}, and Longformer~\cite{beltagy2020longformer}.
These use extra global tokens that gather long-range information as a model memory.

\myparagraph{Vision Transformers} Recent progress has demonstrated that variants of Transformers~\cite{vaswani2017attention} can also be competitive alternatives to CNNs and achieve promising results on different vision tasks.
In addition to image classification, Transformers have also been applied to various vision tasks, including object detection~\cite{carion2020end,dai2021up,zheng2021end,zhu2021deformable},
image generation~\cite{chen2020generative,parmar2018image}, and video processing~\cite{zeng2020learning,zhou2018end}.
Vision Transformer (ViT)~\cite{dosovitskiy2020animage} splits images as small patches and treats the patches as the input word tokens.
ViT shows better performance than CNN-type models with sufficient extensive training data.
DeiT~\cite{tourvron2021deit} incorporates knowledge distillation techniques into ViT training so that we can train a competitive Transformer using only ImageNet-1k~\cite{deng2009imagenet}.
LV-ViT~\cite{jiang2021all} further improves the performance of ViT by introducing a new training objective named token labelling.
Most of these methods have quadratic complexity of self-attention with respect to the input image size.

\myparagraph{Efficient Vision Transformers} There is a thrust to model long sequences of image patches at much higher resolutions.
Recent works such as Pyramid Vision Transformer (PVT)~\cite{wang2021pyramid}, Swin-Transformer~\cite{liu2021swin}, T2T-ViT~\cite{yuan2021tokens},
and Vision Longformer (ViL)~\cite{zhang2021multi} apply transformer layers on different resolution scales by stacking a pyramid of ViTs to form a multi-scale architecture.
To achieve linear complexity, Swin-Transformer~\cite{liu2021swin} uses shifted local window attention.
Vision Longformer~\cite{zhang2021multi} adapts the local attention pattern with the global memory tokens from Longformer~\cite{beltagy2020longformer}.
TimeSformer~\cite{bertasius2021space} applies multiple attentions, each along a single axis of the input video. %
Those methods all leverage fixed, predefined attention patterns to reduce the quadratic cost.
In contrast, our method generates sparse dynamic attention patterns based on the input content.
Another group of works reduce the number of tokens by pruning~\cite{feng2022evit,rao2021dynamicvit,tang2022patch}, or merging tokens~\cite{renggli2022learning,ryoo2021tokenlearner,zeng2022not}.
Recent work, DynamicViT~\cite{rao2021dynamicvit} and EViT~\cite{feng2022evit} study unstructured token sparsification by gradually dropping tokens in the inference of ViTs~\cite{dosovitskiy2020animage}.
However, quadratic attention cost remains in early layers where input tokens cannot be largely sparsified.
Our method instead prunes connectivities at every layer, allowing complexity savings at early layers.

\section{Method}

\begin{figure*}[t]
	\centering
	\includegraphics[width=\linewidth]{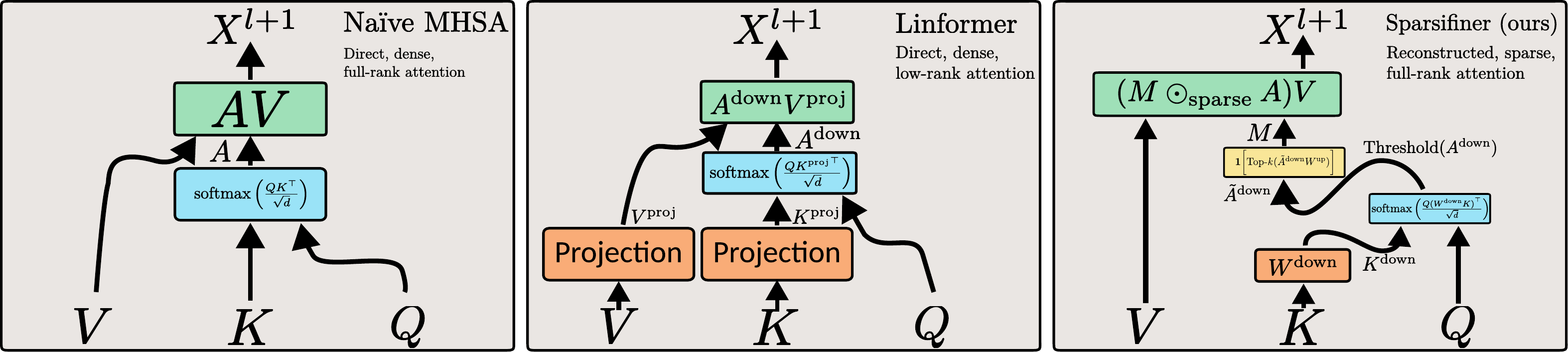}
	\caption{\label{fig:naive-linformer-sparsifiner-mhsa}Single head comparison of the MHSA module for
	na\"{\i}ve MHSA~\cite{vaswani2017attention},
	Linformer~\cite{wang2020linformer}, and Sparsifiner.
	\textbf{Na\"{\i}ve MHSA} incurs quadratic~$O(n^2)$ complexity in the number of
	tokens~$n$.
	\textbf{Linformer} reduces the complexity to linear~$O(n n_\text{down})$ by using a
	projection of the key and value matrices to projected
	key~$K^\text{proj}\in\mathbb{R}^{n_\text{down}\times d}$ and
	value~$V^\text{proj}\in\mathbb{R}^{n_\text{down}\times d}$ matrices in a
	low-rank approximation of the attention matrix.
	\textbf{Sparsifiner}'s key insight is to use the low-rank approximation to learn a
	sparse connectivity mask~$M\in\mathbb{R}^{n\times n}$ and sparse
	up-projection basis~$W^\text{up}$.
	Using sparse matrix multiplication, Sparsifiner reduces overall MHSA FLOPs
	relative to Linformer without restricting the attention matrix to be low
	rank.
	Note that in the rightmost column only (Sparsifiner), the attention
	matrix~$A$ is not explicitly constructed, and rather is used to represent
	sparse attention reconstruction (Eq.~\ref{eq:sparse-full-rank-attention-odot}).}
\end{figure*}

\noindent
Our proposed method to learn sparse attention patterns, Sparsifiner, consists
of a normal ViT~\cite{dosovitskiy2020animage} as the backbone with sparse
attention modules at each layer.
Our sparse attention module consists of a connectivity mask predictor and a sparse multi-head self-attention (MHSA) module.
In both training and inference, we generate a sparse connectivity mask by restricting the number of connections predicted by the mask predictor according to a hyperparameter budget size $B$.
Following this, a sparse MHSA module is used to perform sparse attention based on the connectivity mask.
The sparse MHSA module implements an efficient computation using a sparse
element-wise product between the full attention map and the sparse connectivity
mask to produce a sparse reconstructed attention map.
Then, a sparse-dense attention-value product between the sparse reconstructed
attention map and the value matrix produces the output of the sparse MHSA
module.

For clarity, in the following we describe MHSA for a single attention head only.
In practice, we apply the proposed method to each attention head in a ViT.
We concatenate the resulting output values from all attention heads and feed them to
a linear layer to produce the input to the next transformer
layer~\cite{vaswani2017attention}.

\myparagraph{ViT Architecture and Na\"{\i}ve MHSA} We base our method on the existing ViT model architecture~\cite{dosovitskiy2020animage} and na\"{\i}ve implementation of MHSA~\cite{vaswani2017attention}.  A ViT first tokenizes an input image~$I\in\mathbb{R}^{h\times w\times 3}$ into a set of~$n$ tokens~$X\in\mathbb{R}^{n\times d}$, each with dimension~$d$. Each token consists of a patch embedding,
retrieved via linear projection of the non-overlapping image patches,
and a positional encoding. The resulting sequence of tokens is then fed into MHSA modules to compute the
attention matrix~$A\in\mathbb{R}^{n\times n}$ as the product of
query~$Q\in\mathbb{R}^{n\times d} = X^l W^Q$ and
key~$K\in\mathbb{R}^{n\times d} = X^l W^K$ matrices, where the learned projection matrices~$W^Q\in\mathbb{R}^{d\times d}$
and~$W^K\in\mathbb{R}^{d\times d}$ compute query and key as projections of the
input $X^l\in\mathbb{R}^{n\times d}$ to layer~$l$.
Na\"{\i}ve MHSA then computes the attention
matrix~$A$ as the softmax of outer product of query and key matrices as shown in the left part of Fig.~\ref{fig:naive-linformer-sparsifiner-mhsa}.

\myparagraph{Connectivity Mask Predictor} To enable instance-dependent and meaningful attention patterns while limiting the number of connections, we train a connectivity mask predictor and achieve sparsity by thresholding.
Specifically, we first compute the low-rank approximation
$A^\text{down}\in\mathbb{R}^{n \times n_\text{down}}$ of the attention
matrix~$A$
\begin{equation}
	A^\text{down} = \text{softmax}\bigg(\frac{Q{(W^\text{down} K)}^\top}{\sqrt{d}}\bigg),
	\label{eq:low-rank-attention}
\end{equation}
which we sparsify by thresholding:
\begin{equation}
	\tilde{A}^\text{down}_{ij} = \begin{cases}
		A^\text{down}_{ij} & \text{if } A^\text{down}_{ij} > \tau \\
		0                  & \text{otherwise}
	\end{cases}.
	\label{eq:sparsify-attention}
\end{equation}
In the low-rank attention computation (Eq.~\ref{eq:low-rank-attention}),
we first down-project the token dimension of key matrix~$K$ to a lower
dimension~$n_\text{down}$ using a learned projection
matrix~$W^\text{down}\in\mathbb{R}^{n_\text{down}\times n}$.
Then, a low-rank approximation of the attention matrix is computed from the
outer product of query and down-projected key matrices.
Note that in the low-rank attention sparsification
(Eq.~\ref{eq:sparsify-attention}), with a sparse matrix representation we need
not explicitly store the zeros.

Next, the connectivity mask predictor (Eq.~\ref{eq:connectivity-mask-predictor})
performs a sparse matrix multiplication of a sparse up-projection
matrix~$W^\text{up}\in\mathbb{R}^{n_\text{down}\times n}$ followed by
binarization.
This produces an up-projected sparse connectivity mask:
\begin{equation}
	M = \mathbf{1}\bigg[\text{Top-}k(\tilde{A}^\text{down}W^\text{up}) \bigg].
	\label{eq:connectivity-mask-predictor}
\end{equation}
Here,~$\tilde{A}^\text{down}W^\text{up}$ denotes sparse-sparse matrix
multiplication, which is efficiently computed.
Our key insight is that the post-softmax low-rank attention matrix
(Eq.~\ref{eq:low-rank-attention}) should naturally be sparse.
We show an example in Fig.~\ref{fig:sparse-basis-coef-visualization}.

We apply top-$k$ on the up-projected sparse attention
matrix~$\tilde{A}^\text{down}W^\text{up}$, which is the attention connectivity
score map.
$k$ is set to the budget size $B$.
We discard zero values and binarize to produce a
sparse low-rank connectivity mask~$M\in\mathbb{R}^{n\times n}$.
We indicate binarization by the indicator function~$\mathbf{1}[\cdot]$ in the
connectivity mask predictor (Eq.~\ref{eq:connectivity-mask-predictor}).

\myparagraph{Sparse MHSA}
In Fig.~\ref{fig:naive-linformer-sparsifiner-mhsa}, we compare our method to na\"{\i}ve MHSA~\cite{vaswani2017attention} and
Linformer~\cite{wang2020linformer} in a single head example.
In our method, guided by the sparse
connectivity mask~$M$, we compute only the nonzero elements of the
sparse full-rank attention matrix~$\tilde{A}$.
In order to ensure computational efficiency, we want to have both a sparse up-projection
and a sparse low-rank attention matrix.
This is equivalent to reconstructing the sparse attention matrix~$\tilde{A}$ as an
affine combination over a set of sparse basis vectors using a sparse coefficient vector:
\begin{equation}
	\tilde{A}_{ij} = {\text{softmax}\bigg({\frac{Q K^\top}{\sqrt{d}}}\bigg)}_{ij} \quad\textrm{iff}\quad {M}_{ij} = 1.
	\label{eq:sparse-full-rank-attention}
\end{equation}
Another way of formulating the sparse full-rank attention matrix is as a sparse
element-wise product of the sparse connectivity mask~$M$ with the full-rank
attention matrix:
\begin{equation}
	\tilde{A} = M \odot_\text{sparse} A.
	\label{eq:sparse-full-rank-attention-odot}
\end{equation}
Here, $\odot_\text{sparse}$ is the sparse element-wise product operator, which
skips multiplications by zero.
Therefore, computing the sparse full-rank attention matrix~$\tilde{A}$
(Eq.~\ref{eq:sparse-full-rank-attention}) costs only as many FLOPs as there are
nonzero elements in the connectivity mask~$M$.
In particular, computing the sparse full-rank attention matrix costs less than
the~$O(n^2d)$ required by na\"{\i}ve MHSA\@.

Finally, Sparsifiner computes a sparse attention-value product using the
sparse full-rank attention matrix~$\tilde{A}$ and the value matrix~$V$:
\begin{equation}
	X^{l + 1} = \tilde{A}V.
	\label{eq:sparse-attention-value-product}
\end{equation}
By computing the sparse full-rank attention matrix~$\tilde{A}$
(Eq.~\ref{eq:sparse-full-rank-attention}) guided by the sparse connectivity
mask, and then computing the sparse attention-value product, we remove
the~$O(n^2d)$ complexity required by the na\"{\i}ve MHSA operation.
Instead, the sparse MHSA operation in Sparsifiner performs a number of
operations proportional to the number of nonzero elements in the connectivity
mask~$M$.

\myparagraph{Objective functions}
The training of Sparsifiner includes training the attention connectivity predictor modules and fine-tuning the backbone to make it adapt to sparse attention.
We adopt the standard cross-entropy loss:
\begin{equation}
	\mathcal{L}_{\text{cls}} = \text{CrossEntropy}(\textbf{y}^\text{pred}, \textbf{y})
	\label{eq:cross-entropy-loss}
\end{equation}
where $\textbf{y}^\text{pred}$ is the predicted class distribution and $\textbf{y}$ is the ground-truth class distribution.

To minimize the influence on performance of the attention sparsification process, we use a pre-trained backbone model as a teacher within a knowledge distillation framework.
First, we make the tokens at the last layer close to the ones of the teacher model, where $\textbf{x}$ and $\textbf{x}^{\text{teach}}$
are the tokens after the last block of the Sparsifiner and the teacher model, respectively.

\begin{equation}
	\mathcal{L}_{\text{distill}}^\text{token} = \text{MSE}(\textbf{x}, \textbf{x}^{\text{teach}}).
	\label{eq:token-distill-loss}
\end{equation}
Second, we minimize the difference of Sparsifiner and the teacher model's predictions via KL divergence:
\begin{equation}
	\mathcal{L}_{\text{distill}}^\text{cls} = \text{KL}(\textbf{y}^\text{pred} || \textbf{y}^\text{teach}).
	\label{eq:cls-distill-loss}
\end{equation}

Third, we want the connectivity score map generated by the connectivity mask predictor to be a good low-rank
approximation of the teacher attention, which can be viewed as knowledge distillation of the attention map.
We minimize the Euclidean distance between them:
\begin{equation}
	\mathcal{L}_{\text{distill}}^\text{attn} = \text{MSE}(\tilde{A}^\text{down}W^\text{up}, A^\text{teach}).
	\label{eq:attn-reconstruct-loss}
\end{equation}

Finally, to enforce the sparsity of the up-projection matrix, we use the $L_2$ regularization.
We tried $L_1$ regularization but found that $L_2$ gives better training convergence with sufficient sparsity in practice.
\begin{equation}
	\mathcal{L}_{\text{spa}} = \sum_{i} (w^\text{up}_{i})^2
	\label{eq:sparse-basis-loss}
\end{equation}

The full training objective combines all objectives:
\begin{equation}
	\mathcal{L}  = \mathcal{L}_{\text{cls}} +
	\lambda_{\text{distill}}^\text{token}\mathcal{L}_{\text{distill}}^\text{token} +
	\lambda_{\text{distill}}^\text{cls}\mathcal{L}_{\text{distill}}^\text{cls} +
	\lambda_{\text{distill}}^\text{attn}\mathcal{L}_{\text{distill}}^\text{attn} +
	\lambda_{\text{spa}}\mathcal{L}_{\text{spa}}
	\label{eq:total-loss}
\end{equation}
Where we set the weight decay as $0.05$ in the optimizer instead of directly adding $\lambda_{\text{spa}}\mathcal{L}_{\text{spa}}$ to the objective.

\begin{figure*}[t]
	\centering
	\includegraphics[width=\linewidth]{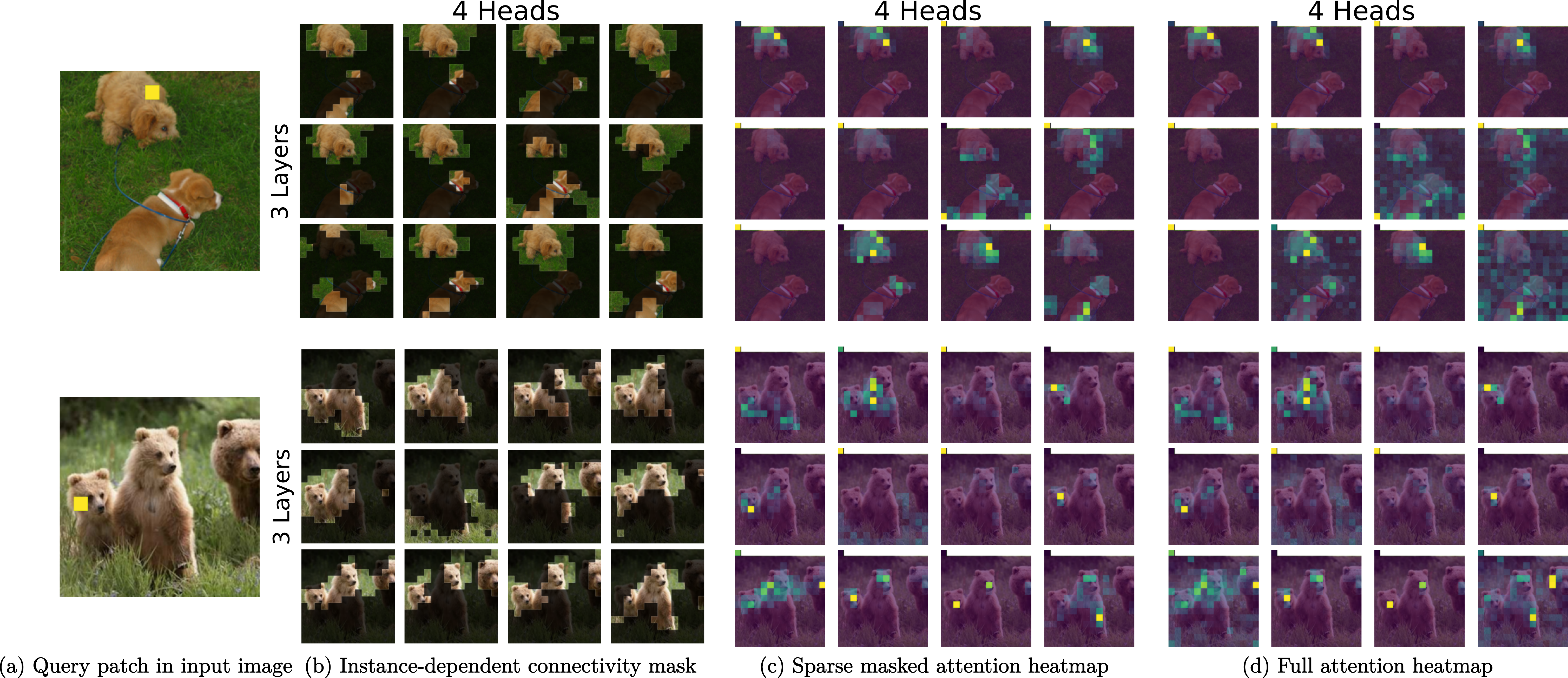}
	\caption{\label{fig:attn-map-vis-bear-dogs}Visualization of connectivity mask (b)
		with sparse (c) and full (d) attention maps for a given query patch (a).
		In the heatmaps, the blue darker color indicates lower, and yellow brighter
		color indicates higher attention value.
		Here we visualize the attention maps for only 3 layers and 4 heads of the
		ViT\@.
		For the dog image (top) we visualize layers 3--5, while for the bear image
		(bottom) we visualize layers 6--8.
		We observe that in earlier layers the attention map focuses more on
		positional information such as nearby tokens, while in later layers semantic
		relations with distant tokens are more important.
		For each query patch indicated by a yellow square in the input image,
		Sparsifiner predicts a sparse connectivity mask using a low-rank
		approximation to full attention.
		Using the sparse connectivity mask, Sparsifiner efficiently computes a
		sparse full-rank attention matrix.
		By comparison with the rightmost full attention, sparse attention retains all of the
		most salient relations with the given query patch, while discarding redundant
		or noisy information in the rest of the image.
	}
\end{figure*}

\section{Experiments and Results}%
\label{sec:experiments}

\noindent
\myparagraph{Implementation details} We train all of the models on the ImageNet dataset~\cite{deng2009imagenet}.
By default, the connectivity mask predictor module is incorporated into every layer of DeiT-S~\cite{tourvron2021deit} and LV-ViT-S~\cite{jiang2021all}.
In all of our experiments, we set the reduced dimension~$n_\text{down}$ to~\num{32} and $\tau$ to~\num{0.05} which ensures 87\% sparsity ratio of the basis coefficient.
The attention budget~$B$ is in the range~$(0, \text{number of tokens}]$.
Budget~$B$ is directly determined by the attention keep rate in~$(0, 1]$ as
the ceiling of the keep rate multiplied by the total number of tokens.

We follow most of the training techniques used in DeiT-S and LV-ViT-S.
We use pre-trained ViT models to initialize the backbone models.
To improve speed of convergence, we propose a two-phase training strategy.
In the first phase, we freeze the backbone model and train the connectivity mask predictor module with attention distillation loss and L2 regularization only.
Specifically, we set $\lambda_{\text{distill}}^\text{token} = 0.0$, $\lambda_{\text{distill}}^\text{cls} = 0.0$, $\lambda_{\text{distill}}^\text{attn} = 1.0$
and we also apply a threshold 1e-2 on basis $W^\text{up}$ to ensure 90\% sparsity.
We found that this setting helps the connectivity mask predictor to learn $W^\text{up}$ quickly and loss converges within 5 epochs.
In the second phase, we jointly train the backbone model and the connectivity mask predictor module for another 40 epochs.
we set $\lambda_{\text{distill}}^\text{token} = 0.5$, $\lambda_{\text{distill}}^\text{cls} = 0.5$, $\lambda_{\text{distill}}^\text{attn} = 0.0$.
More details can be found in the supplementary material.

\myparagraph{Sparse connectivities and attention visualization} In order to
qualitatively investigate the quality of Sparsifiner's sparse attention approximation,
we visualize its connectivity mask and sparse reconstructed attention map
(Fig.~\ref{fig:attn-map-vis-bear-dogs}).
We show the original input image and the connectivity mask of the query patch,
where the dark regions represent tokens that are not attended to by the query
patch token.
For each attention head, Sparsifiner generates a corresponding connectivity mask.
We find that the connectivity mask acts as a region proposal mechanism, which
allows different attention heads to locate different informative tokens and
gather diverse semantic information.
Furthermore, we visualize the sparse attention map efficiently generated using
the connectivity mask and compare it with the full attention map.
We find that the sparse attention map retains all of the highest connectivity
values, while discarding lower connectivity values.
Hence the visualizations show that Sparsifiner retains the most salient
relations for a given token, while discarding noisy background relations.

\begin{figure}[t]
	\centering
	\includegraphics[width=\linewidth]{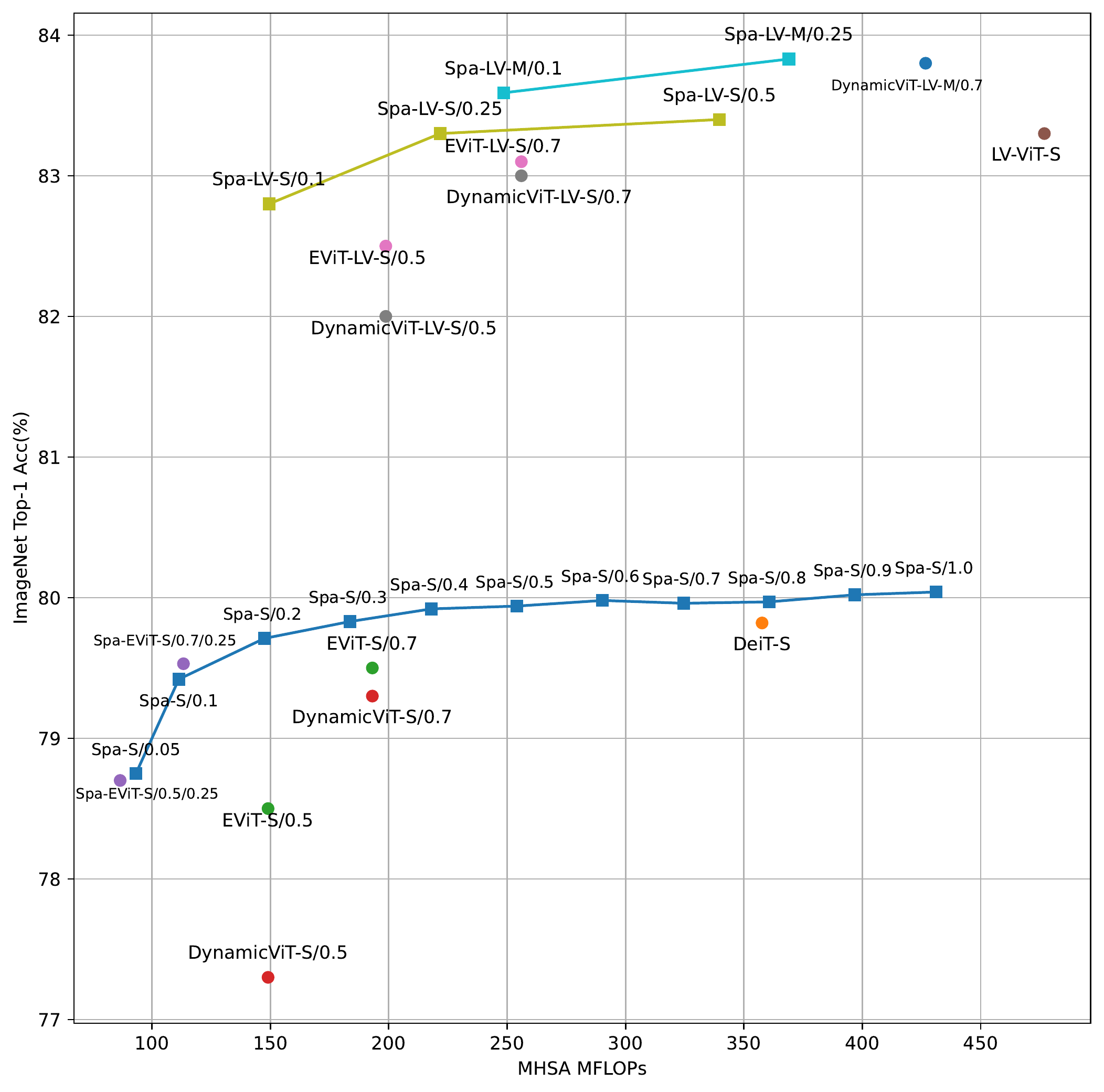}
	\caption{\label{fig:mhsa-mflops-vs-accuracy}
		MHSA computation (FLOPs) and top-1 accuracy trade-offs on ImageNet. We compare Sparsifiner with the
		state-of-the-art token pruning methods. Sparsifiner achieves superior trade-offs compared to the baseline.
		We also report MHSA FLOPs and top-1 accuracy for Sparsifiner-S under varying attention keep rate.
	}
\end{figure}

\begin{table}[t]\centering
	{\small
		\begin{tabular}{@{}lcccc@{}}
			\toprule
			Model                                & \begin{tabular}{@{}c@{}}Tok. \\ keep \\ rate\end{tabular} & \begin{tabular}{@{}c@{}}Att. \\ keep \\ rate\end{tabular} & \begin{tabular}{@{}c@{}}MHSA \\ (MFLOPs)\end{tabular}           & \begin{tabular}{@{}c@{}}Top-1 \\ Acc. \\ (\%)\end{tabular} \\
			\midrule
			DeiT-S~\cite{tourvron2021deit}       & \num{1.0}                  & \num{1.0}                  & \num{357.7}                          & \num{79.8}                 \\
			\midrule
			EViT~\cite{feng2022evit}             & \num{0.7}                  & \num{1.0}                  & \num{193.1} (\num{-46}\%)            & $\mathbf{79.5}$            \\
			DynamicViT~\cite{rao2021dynamicvit}  & \num{0.7}                  & \num{1.0}                  & \num{193.1} (\num{-46}\%)            & \num{79.3}                 \\
			\textbf{Sparsif-EViT (ours)}         & \num{0.7}                  & \num{0.25}                 & \num{113.3}  (\num{-68}\%)           & $\mathbf{79.5}$            \\
			\textbf{Sparsifiner (ours)}          & \num{0.7}                  & \num{0.25}                 & \num{113.3}  (\num{-68}\%)           & \num{79.3}                 \\
			\midrule
			EViT~\cite{feng2022evit}             & \num{0.5}                  & \num{1.0}                  & \num{149.1} (\num{-58}\%)            & \num{78.5}                 \\
			DynamicViT~\cite{rao2021dynamicvit}  & \num{0.5}                  & \num{1.0}                  & \num{149.1} (\num{-58}\%)            & \num{77.3}                 \\
			\textbf{Sparsif-EViT (ours)}         & \num{0.5}                  & \num{0.25}                 & $\mathbf{86.6}$ ($\mathbf{-76}$\%)   & \num{78.7}                 \\
			\textbf{Sparsifiner (ours)}          & \num{0.5}                  & \num{0.25}                 & $\mathbf{86.6}$ ($\mathbf{-76}$\%)   & \num{78.4}                 \\
			\midrule
			LV-ViT-S~\cite{jiang2021all}         & \num{1.0}                  & \num{1.0}                  & \num{476.9}                          & \num{83.3}                 \\
			\midrule
			EViT-LV-S~\cite{feng2022evit}        & \num{0.7}                  & \num{1.0}                  & \num{256.0} (\num{-46}\%)            & \num{83.0}                 \\
			EViT-LV-S~\cite{feng2022evit}        & \num{0.5}                  & \num{1.0}                  & \num{198.8} (\num{-58}\%)            & \num{82.5}                 \\
			DynViT-LV-S~\cite{rao2021dynamicvit} & \num{0.7}                  & \num{1.0}                  & \num{256.0} (\num{-46}\%)            & \num{83.0}                 \\
			DynViT-LV-S~\cite{rao2021dynamicvit} & \num{0.5}                  & \num{1.0}                  & \num{198.8} (\num{-58}\%)            & \num{82.0}                 \\
			\textbf{Sparsif-LV-S (ours)}         & \num{1.0}                  & \num{0.5}                  & \num{339.7}  (\num{-29}\%)           & $\mathbf{83.4}$            \\
			\textbf{Sparsif-LV-S (ours)}         & \num{1.0}                  & \num{0.25}                 & \num{221.7}  (\num{-54}\%)           & \num{83.3}                 \\
			\textbf{Sparsif-LV-S (ours)}         & \num{1.0}                  & \num{0.1}                  & $\mathbf{149.5}$  ($\mathbf{-69}$\%) & \num{82.8}                 \\
			\bottomrule
		\end{tabular}
		\caption{\label{tab:token-pruning}Comparison with token pruning methods on
			DeiT-S~\cite{tourvron2021deit} and LV-ViT-S~\cite{jiang2021all} base models.
			Token pruning methods such as EViT~\cite{feng2022evit} and
			DynamicViT~\cite{rao2021dynamicvit} prune tokens at fixed layers.
			We show that token pruning methods combine with Sparsifiner's sparse
			attention connectivities to produce a complementary effect.
			Sparsifiner combined with EViT~\cite{feng2022evit} achieves a~\num{68}\% reduction in FLOPs
			compared with the DeiT-S~\cite{tourvron2021deit} baseline, while maintaining a top-1 accuracy
			of~\num{79.5}\%.
			Hence Sparsifiner achieves the same top-1 accuracy as EViT~\cite{feng2022evit} with
			significantly better MHSA FLOPs reduction.
			The input resolution is~$224\times224$.
		}
	}
\end{table}

\myparagraph{Comparison with token pruning}
We train and evaluate Sparsifiner on ImageNet and compare to state-of-the-art
token pruning baselines (Tab.~\ref{tab:token-pruning}).
Since our research question addresses the problem of reducing MHSA complexity,
we report trade-offs between top-1 accuracy on ImageNet and computation in
terms of MHSA FLOPs.
We compare Sparsifiner against baselines by adjusting two hyperparameters:
token and attention keep rate.
The token keep rate is the fraction of tokens kept in the network at
predetermined layers where pruning occurs, which we set according to
established token pruning baselines~\cite{feng2022evit,rao2021dynamicvit}.
The attention keep rate is the fraction of attention connectivities at any
given MHSA layer, as determined by the connectivity mask predictor
(Eq.~\ref{eq:connectivity-mask-predictor}).
Hence, varying the attention keep rate reduces FLOPs without necessitating
removal of tokens as in token pruning.
But both techniques can be combined to achieve complementary effects.

To provide a variety of comparisons we experiment with adding token pruning and
Sparsifiner to two common baseline ViT models: DeiT~\cite{tourvron2021deit} and
LV-ViT~\cite{jiang2021all}.
On both models, Sparsifiner achieves significant computation saving while
maintaining a relatively modest drop in top-1 accuracy.
For example, LV-ViT-S~\cite{jiang2021all} trained with Sparsifiner with an attention keep rate
of~\num{0.25} reduces the MHSA FLOPs by~\num{53.5}\% while maintaining the
top-1 accuracy of the baseline LV-ViT-S model on ImageNet.
When used in combination with token pruning, Sparsifiner achieves an even
superior reduction in MHSA FLOPs while maintaining comparable top-1
accuracy to EViT, and superior top-1 accuracy to DynamicViT\@.

\begin{table}[t]\centering
	{\small
		\begin{tabular}{@{}llccc@{}}
			\toprule
			Att. keep rate                             & Att. num. & \begin{tabular}{@{}c@{}}MHSA \\ (MFLOPs)\end{tabular} & \begin{tabular}{@{}c@{}}Top-1 \\ Acc (\%)\end{tabular} \\
			\midrule
			\num{1.0} (DeiT-S~\cite{tourvron2021deit}) & \num{197} & \num{357.7}                & \num{79.82}                \\
			\midrule
			\num{0.9}                                  & \num{178} & \num{396.8}                & \num{80.02}                \\
			\num{0.8}                                  & \num{158} & \num{360.6}                & \num{79.97}                \\
			\num{0.7}                                  & \num{138} & \num{324.6} (\num{-9}\%)   & \num{79.96}                \\
			\num{0.6}                                  & \num{119} & \num{290.3} (\num{-19}\%)  & \num{79.98}                \\
			\num{0.5}                                  & \num{99}  & \num{254.2} (\num{-29}\%)  & \num{79.94}                \\
			\num{0.4}                                  & \num{79}  & \num{218.0} (\num{-39}\%)  & \num{79.92}                \\
			\num{0.3}                                  & \num{60}  & \num{183.6} (\num{-49}\%)  & \num{79.83}                \\
			\num{0.2}                                  & \num{40}  & \num{147.5} (\num{-59}\%)  & \num{79.71}                \\
			\num{0.1}                                  & \num{20}  & \num{111.4} (\num{-69}\%)  & \num{79.42}                \\
			\num{0.05}                                 & \num{10}  & \num{93.3}  (\num{-74}\%)  & \num{78.75}                \\
			\num{0.01}                                 & \num{2}   & \num{78.9}  (\num{-78}\%)  & \num{73.03}                \\
			\bottomrule
		\end{tabular}
		\caption{\label{tab:attention-budget}Effect of attention budget on FLOPs and top-1 accuracy.
			Here the ``keep rate'' refers to the number of attention connectivities retained at each layer.
			All other attention connectivities in the sparse full-rank attention
			matrix (Eq.~\ref{eq:sparse-full-rank-attention}) are set to zero.
			When keeping only~\num{10} attention connectivities, Sparsifiner produces
			a top-1 accuracy reduced by only~\num{1.0}\% compared to the
			full-attention baseline DeiT-S~\cite{tourvron2021deit}, but with a~\num{73.9}\% reduction in
			FLOPs.
			The input resolution is~$224\times224$.
		}
	}
\end{table}

\myparagraph{Varying MHSA attention budget}
We varied the attention budget of MHSA in order to investigate the tradeoff
between MHSA FLOPs and top-1 accuracy for Sparsifiner-S
(Tab.~\ref{tab:attention-budget}).
The results evaluated on ImageNet show that Sparsifiner-S produces a superior
Pareto frontier compared with previous approaches
(Fig.~\ref{fig:mhsa-mflops-vs-accuracy}).
In particular, Sparsifiner-S models with attention budgets of~\num{40} and
above achieved top-1 accuracy within~\num{0.1}\% of the full-rank DeiT-S~\cite{tourvron2021deit} model,
while using~\num{58.8}\% fewer FLOPs in MHSA\@.
Furthermore, Sparsifiner-S models with high attention budgets of~\num{79} and
above achieved superior top-1 accuracy compared with the full-rank DeiT-S~\cite{tourvron2021deit}
model, while using fewer FLOPs in MHSA\@.
This suggests that Sparsifiner's sparse full-rank attention reconstruction
mechanism induces a useful regularization effect that improves model
generalization.

\begin{table}[t]\centering
	{\small
		\begin{tabular}{@{}lcccc@{}}
			\toprule
			Model                  & \begin{tabular}{@{}c@{}}Att. \\ keep \\ rate\end{tabular} & \begin{tabular}{@{}c@{}}MHSA \\ (MFLOPs)\end{tabular}  & \begin{tabular}{@{}c@{}}Overall \\ (GFLOPs)\end{tabular}  & \begin{tabular}{@{}c@{}}Top-1 \\ Acc \\ (\%)\end{tabular} \\
			\midrule
			DeiT-T                 & \num{1.0}                  & \num{1534.1}                & \num{3.58}                  & \num{75.45}                \\
			\midrule
			\textbf{Sparsifiner-T} & \num{0.5}                  & \num{851.0}   (\num{-45}\%) & \num{2.89}   (\num{-19}\%)  & \num{75.45}                \\
			\textbf{Sparsifiner-T} & \num{0.25}                 & \num{452.9}   (\num{-70}\%) & \num{2.49}    (\num{-30}\%) & \num{75.35}                \\
			\textbf{Sparsifiner-T} & \num{0.1}                  & \num{240.5}   (\num{-84}\%) & \num{2.28}    (\num{-36}\%) & \num{74.58}                \\
			\bottomrule
		\end{tabular}
		\caption{\label{tab:deit-s-384}Results on high resolution~$384\times384$ images. We apply Sparsifiner on DeiT-T~\cite{tourvron2021deit} with resolution 384.
			We show that Sparsifiner reduce the MHSA complexity of DeiT-T-384~\cite{tourvron2021deit} by over \num{84}\% with modest accuracy drop.
			Since the number of tokens is quadratic in the resolution, Sparsifiner
			can reduce a larger portion of MHSA complexity on
			high-resolution images.
		}
	}
\end{table}

\myparagraph{Accelerating ViT on high-resolution images} To show the effectiveness of our method on larger input size,
we apply our method to DeiT-T~\cite{tourvron2021deit} with~$384\times 384$ resolution (Tab.~\ref{tab:deit-s-384}).
When dealing with high-resolution images, due to quadratic complexity in the
number of tokens, MHSA becomes increasingly expensive compared to the
feedforward operations.
We reduce the MHSA complexity of the DeiT-T~\cite{tourvron2021deit} model with~$384\times 384$ input by over~\num{80}\% with less than~\num{1}\% accuracy drop.
Our method shows a great potential to accelerate ViT on even higher resolution images where token quantity dominates the model complexity.

\begin{table}[t]\centering
	{\small
		\begin{tabular}{@{}lcc@{}}
			\toprule
			Model                              & MHSA (MFLOPs)            & Top-1 Acc (\%) \\
			\midrule
			Linformer~\cite{wang2020linformer} & \linformermhsamflops{}   & 77.54          \\
			\textbf{Sparsifiner-S (ours)}      & \sparsifinermhsamflops{} & 79.79          \\
			\bottomrule
		\end{tabular}
		\caption{\label{tab:linformer}Comparison of sparse full-attention
			reconstruction with low-rank attention reconstruction.
			Sparsifiner-S achieves a \sparsifinerlinformerimprovement{}\% absolute percentage
			point improvement in top-1 accuracy compared with
			Linformer~\cite{wang2020linformer}.}
	}
\end{table}

\begin{figure}[t]
	\begin{subfigure}[b]{0.235\textwidth}
		\centering
		\includegraphics[width=\textwidth]{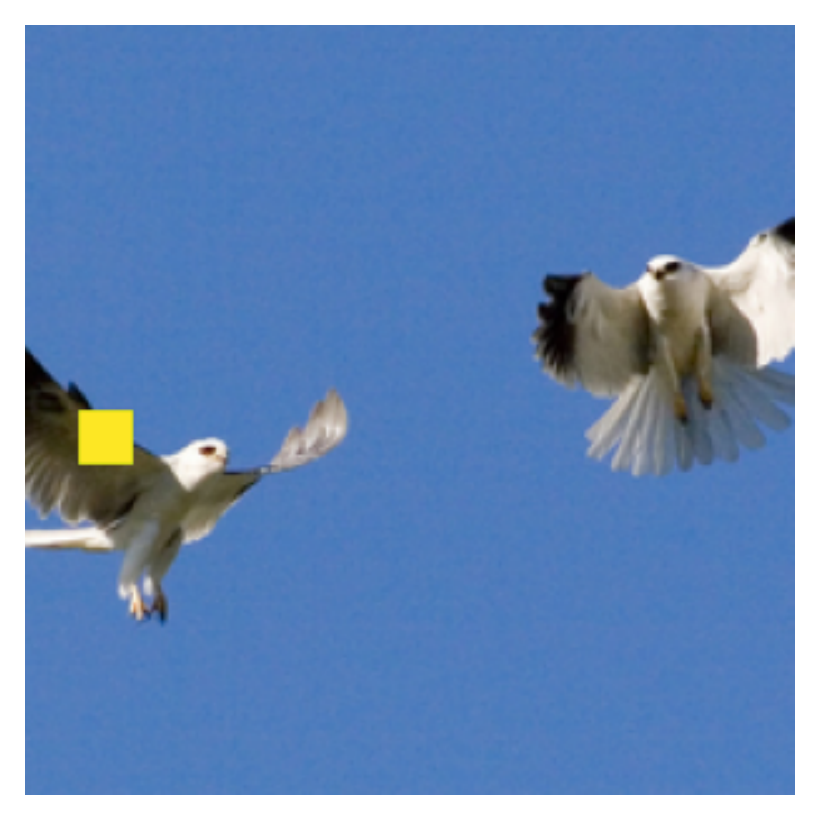}
		\caption{\label{fig:query-patch-1-linformer-low-rank-attention}Query in input image}
	\end{subfigure}
	\hfill
	\begin{subfigure}[b]{0.235\textwidth}
		\centering
		\includegraphics[width=\textwidth]{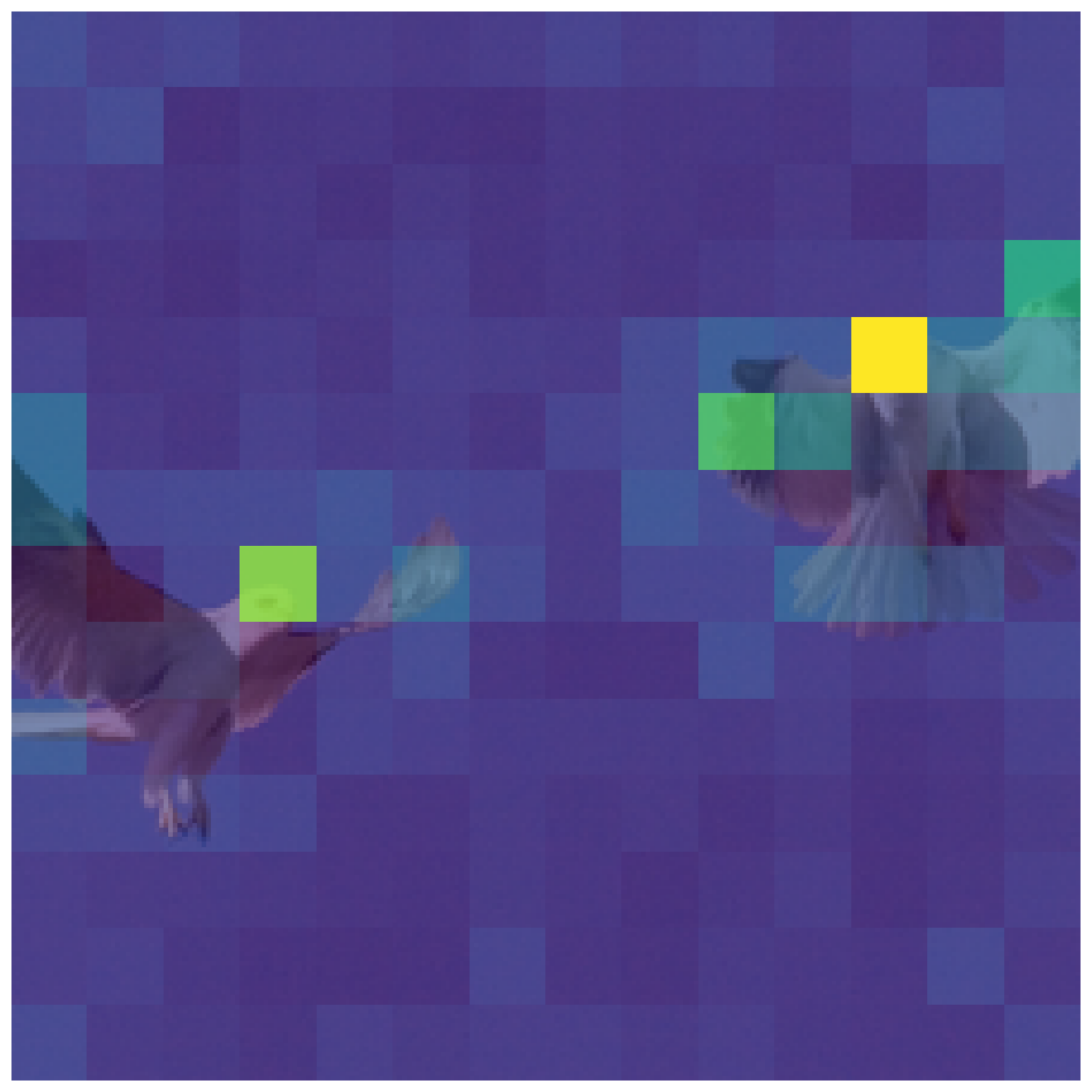}
		\caption{\label{fig:full-attention}Full attention (na\"{\i}ve MHSA)}
	\end{subfigure}
	\hfill
	\begin{subfigure}[b]{0.235\textwidth}
		\centering
		\includegraphics[width=\textwidth]{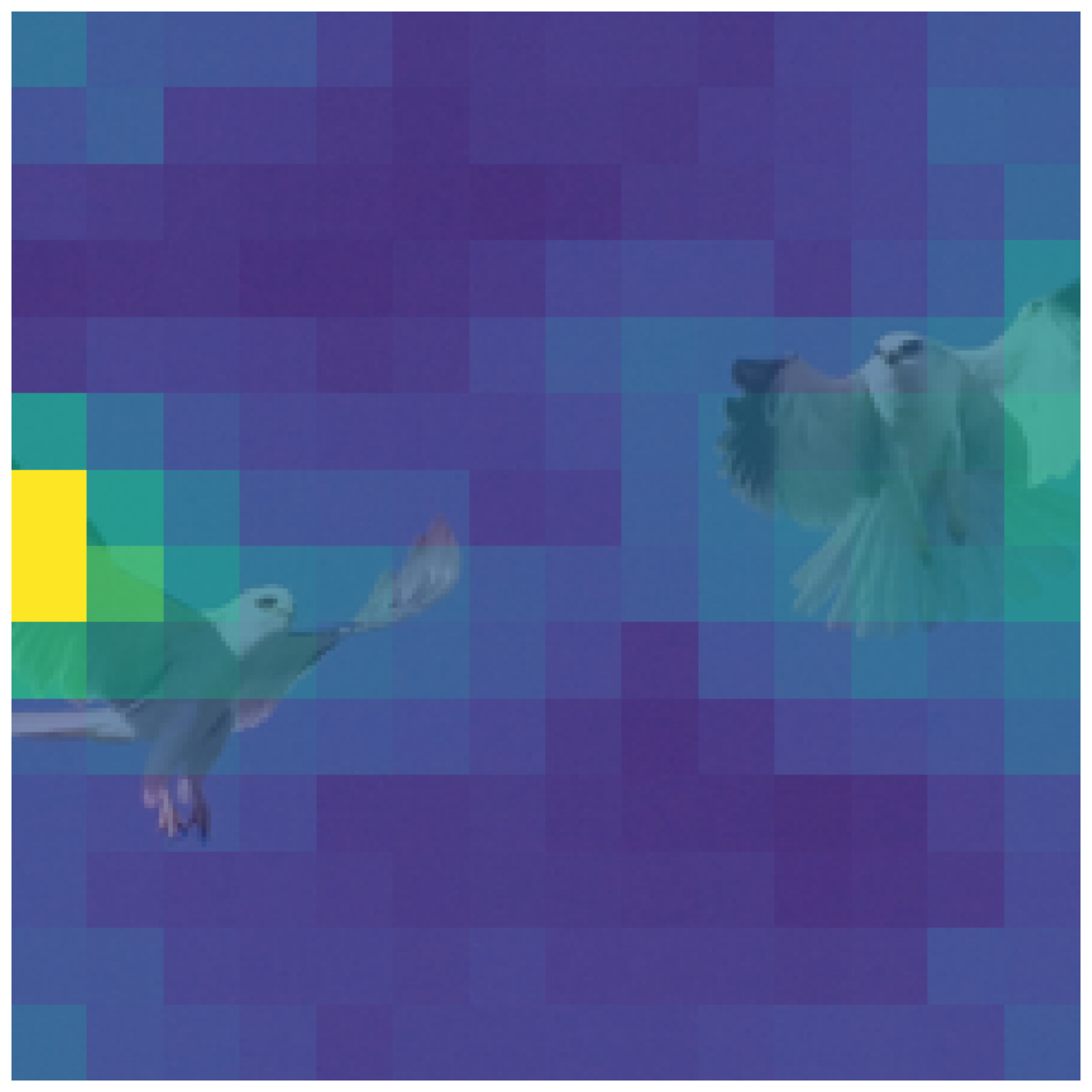}
		\caption{\label{fig:low-rank-attention}Low-rank attention (Linformer)}
	\end{subfigure}
	\hfill
	\begin{subfigure}[b]{0.235\textwidth}
		\centering
		\includegraphics[width=\textwidth]{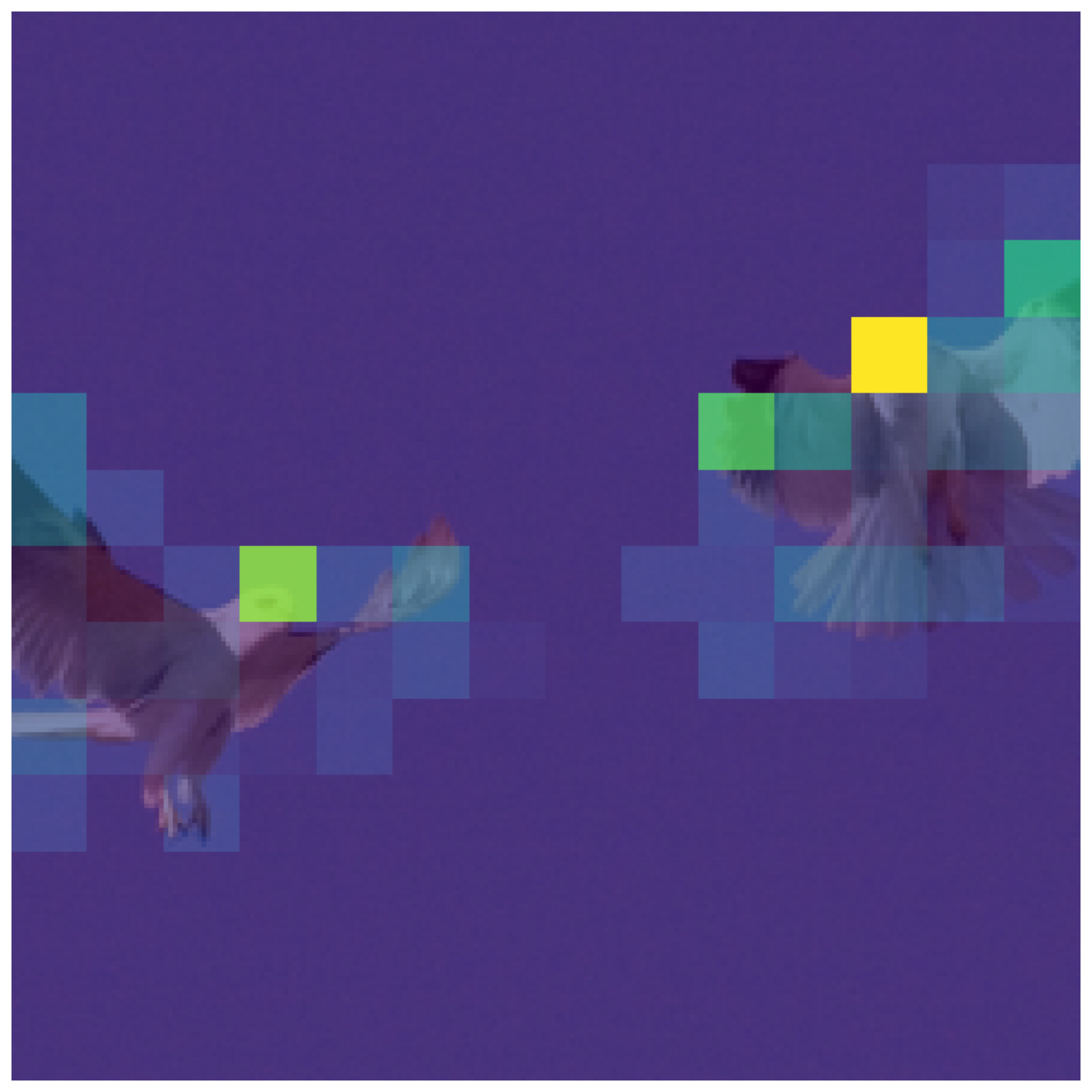}
		\caption{\label{fig:full-rank-attention}Full-rank sparse att. (Sparsifiner)}
	\end{subfigure}
	\caption{\label{fig:full-rank-sparse-attention}Low-rank attention (Linformer) and full-rank sparse attention (Sparsifiner) heatmaps.
		For a given query patch indicated by a yellow square (a), we visualize its
		low-rank attention map (Linformer) (c) and full-rank sparse attention
		map (Sparsifiner) (d).
		Due to discarding the long tail of the attention matrix's eigenspectrum,
		low-rank attention produces a coarse attention map.
		By contrast, full-rank sparse attention bears closer resemblance to
		full attention (b) with low-salience connectivities discarded.}
\end{figure}

\begin{figure*}[t]
	\centering
	\includegraphics[width=0.9\linewidth]{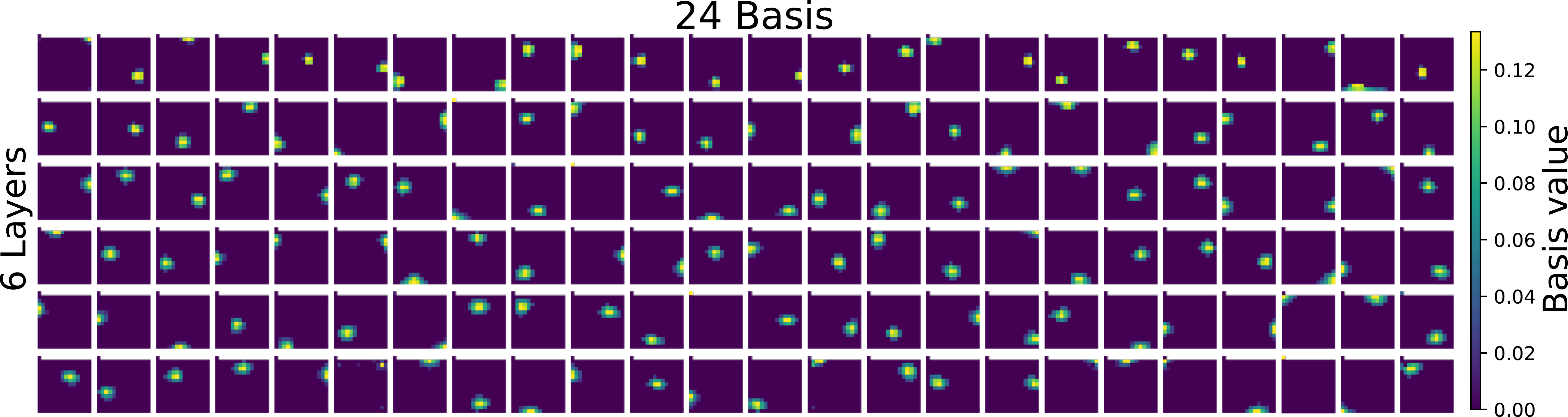}
	\caption{\label{fig:sparse-basis-visualization}Visualization of the
		up-projection matrix~$W^\text{up}$ of the first~\num{6} layers of
		Sparsifiner-S, which we refer to here as a sparse basis.
		We visualize~\num{24} dimensions of the sparse basis.
		Dark blue weights indicate low values, which are pruned after training so
		that only the bright yellow weights are left over.
		Qualitatively, the sparse basis has a high level of sparsity, making sparse attention reconstruction efficient.}
\end{figure*}

\begin{figure}[t]
	\centering
	\includegraphics[width=0.9\linewidth]{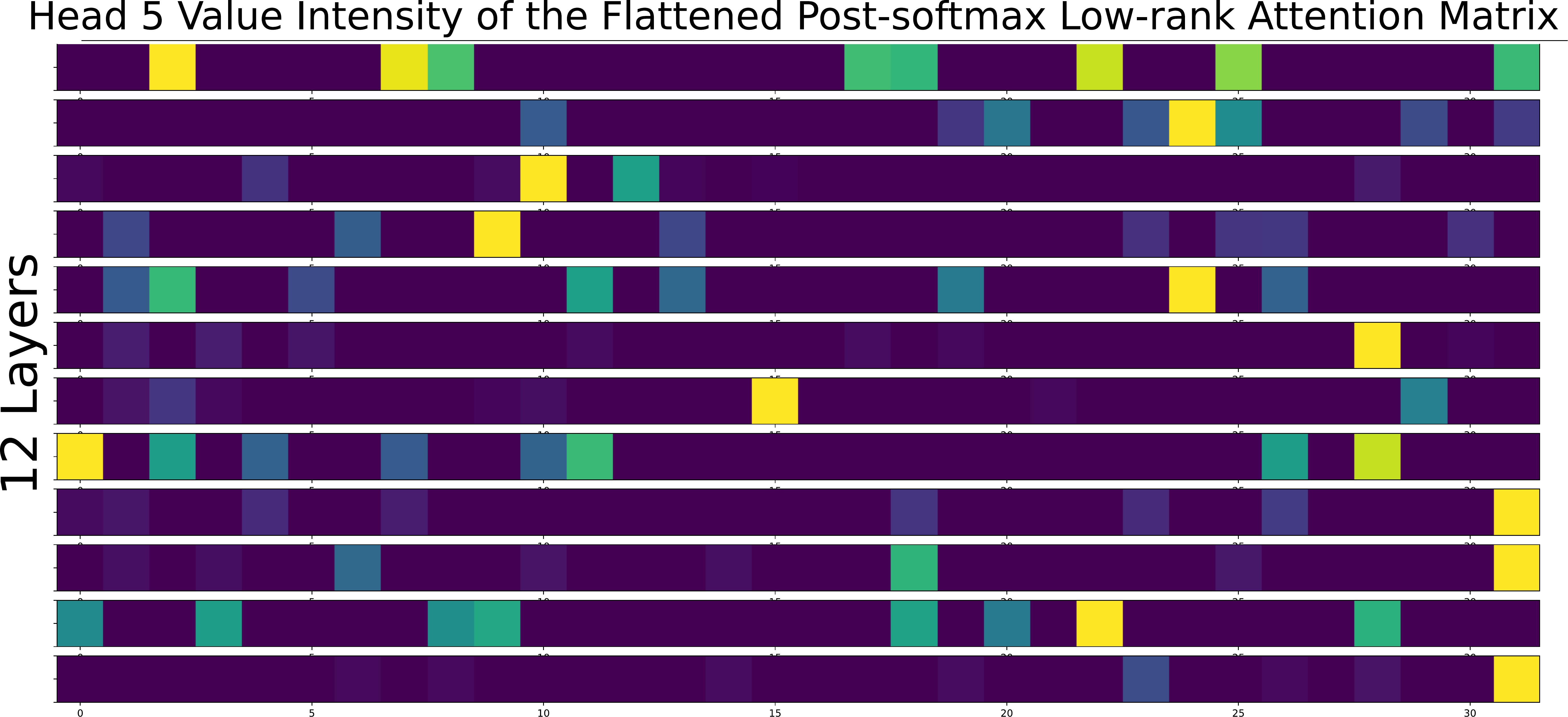}
	\caption{\label{fig:sparse-basis-coef-visualization}Visualization of the sparse basis coefficient of the~\num{5}th
		attention head over~\num{12} layers of Sparsifiner-S.
		Dark blue regions indicate low values that are pruned before sparse
		attention reconstruction during inference, leaving only bright yellow
		coefficients.}
\end{figure}

\myparagraph{Low-rank: connectivities or attention?} Our approach raises a
research question: does the utility of the dense low-rank attention matrix come from
its use as a connectivity mask?
Or is it sufficient to directly use the dense low-rank attention matrix,
foregoing the need to reconstruct the sparse full-rank attention matrix, i.e., the Linformer~\cite{wang2020linformer} approach?
We answered this question by comparing the top-1 accuracy of the two approaches
(Tab.~\ref{tab:linformer}).
In this experiment, Sparsifiner-S and Linformer~\cite{wang2020linformer} were trained under identical
settings, differing only in the attention approximation method.
Sparsifiner-S uses a reconstructed sparse full-rank attention matrix, while
Linformer uses the dense low-rank attention matrix directly.
In order to give both models similar representational capacity, we set the low-rank
dimension of Linformer~\cite{wang2020linformer} equal to the sparse attention budget of Sparsifiner-S.
This enforces that the attention-value product of both models' MHSA has the same complexity.

Using the sparse full-rank attention matrix produces a
\sparsifinerlinformerimprovement{}\% absolute percentage point improvement in top-1
accuracy compared with Linformer.
This improvement reinforces the superiority of using the low-rank query-key product as a
connectivity mask, rather than using the low-rank attention matrix directly.
Using the low-rank attention matrix to directly compute the attention-value product with
a down-projected value discards the long tail of the full attention matrix's
eigenspectrum~\cite{wang2020linformer}.
In contrast, using the low-rank query-key product as a connectivity mask reduces
computation by a different mechanism.
By using a low-rank connectivity mask to produce a sparse full-rank attention matrix,
the long-tail of the full attention matrix's eigenspectrum is preserved.
Based on the significant improvement in top-1 accuracy, we conclude that these long-tail
eigenvalues are important for model predictive quality in ViTs.

\myparagraph{Low- and full-rank attention visualization} In order to further
illuminate the qualitative difference between low- and full-rank attention in
ViTs, we also present the masked attention heatmap and the full attention
heatmap of the query patch (Fig.~\ref{fig:full-rank-sparse-attention}).
We show that a connectivity mask can accurately preserve key tokens that are highly related to the query patch and
remove the irrelevant ones. As a result, the masked attention heatmap preserves structure and discards
noise compared with the full attention heatmap. The visualization results also validate that our Sparsifiner
can effectively approximate the full attention ViT.

\myparagraph{Sparse low-rank basis and up-projection matrix visualization} To
demonstrate that the connectivity mask can be computed by sparse-sparse matrix
multiplication, we visualize the up-projection matrix~$W^\text{up}$ of the
first six layers of Sparsifiner (Fig.~\ref{fig:sparse-basis-visualization}).
Because the reconstructed sparse attention matrix is a combination of the
up-projection matrix's weights, we refer to it as a sparse basis.
We show that Sparsifiner naturally learns a sparse basis of local regions
resembling 2D Gaussians.
For a given token, the sparse bases corresponding to object locations with
salient semantic and/or spatial information will activate.
Since the sparse attention reconstruction
(Eq.~\ref{eq:sparse-full-rank-attention-odot}) is a product of the sparse
low-rank attention matrix with the up-projection matrix,
we also visualize the post-softmax low-rank attention matrix.
Here we view the low-rank attention matrix as a sparse coefficient of the
sparse basis (Fig.~\ref{fig:sparse-basis-coef-visualization}).
Qualitatively, the sparse coefficient also exhibits a high degree of sparsity,
further validating the efficiency of the sparse attention reconstruction via
sparse-sparse matrix multiplication.

\section{Conclusions}

\noindent
We presented a novel computationally efficient approach to learn unstructured,
instance-dependent attention in ViTs.
The development of sparse attention mechanisms such as Sparsifiner opens the
door to further research into accelerating sparse ViTs using software-hardware
systems approaches.
Sparsifiner shows the promise of sparse attention for scaling ViTs to larger
and more complex vision tasks.
But software-hardware systems approaches are needed to realize its full
potential.
We hope that our work inspires further research at the intersection of sparse
algorithms for ViTs and software-hardware systems approaches to support those
sparse algorithms.

\par\vfill\par

\clearpage

{\small
	\bibliographystyle{ieee_fullname}
	\bibliography{sparsifiner}
}

\clearpage

\end{document}